\documentclass{article} 
\usepackage{iclr2024_conference,times}

\usepackage{amsmath,amsfonts,bm}

\def\eqref#1{equation~\ref{#1}}

\def\1{\bm{1}}

\DeclareMathAlphabet{\mathsfit}{\encodingdefault}{\sfdefault}{m}{sl}
\SetMathAlphabet{\mathsfit}{bold}{\encodingdefault}{\sfdefault}{bx}{n}

\usepackage{hyperref}
\usepackage{url}

\usepackage{amsmath,amsfonts,bm}
\usepackage{amsthm}

\usepackage{amssymb,mathtools}
\usepackage{bbm}
\usepackage{nccmath}

\usepackage{subcaption}
\usepackage{booktabs}
\usepackage{multirow}
\usepackage{arydshln}

\usepackage{xfrac}
\usepackage{mdframed}
\usepackage{xspace}
\usepackage{bbding}
\usepackage{mathtools}
\usepackage{soul}
\usepackage{pgf}
\usepackage{array}
\usepackage{enumitem}

\usepackage{fontawesome5}

\usepackage{tikz}
\usepackage{tikz-dependency}

\newcommand{\rebuttal}[1]
{{\color{black}{#1}}}
\newcommand{\mcite}[1]{\mbox{\cite{#1}}}

\usepackage[capitalize,nameinlink]{cleveref}
\Crefname{section}{\S\hspace{-1mm}}{\S\hspace{-0.5mm}}
\Crefname{appendix}{App.}{Apps.}

\usepackage{adjustbox}

\usepackage{mdframed}
\definecolor{theoremcolor}{rgb}{0.97, 0.97, 0.97}
\definecolor{examplecolor}{rgb}{1, 1, 1.0}
\mdfsetup{
    innertopmargin=8pt,
    innerbottommargin=8pt,
    leftmargin=4pt,
    rightmargin=4pt,
    backgroundcolor=theoremcolor,
    linewidth=0pt,
}
\usepackage{xfrac}
\usepackage{comment}
\newmdtheoremenv[linewidth=0pt,innerleftmargin=4pt,innerrightmargin=4pt]{definition}{Definition}
\newmdtheoremenv[linewidth=0pt,innerleftmargin=4pt,innerrightmargin=4pt]{proposition}{Proposition}
\newmdtheoremenv[linewidth=0pt,innerleftmargin=0pt,innerrightmargin=0pt,backgroundcolor=examplecolor]{example}{Example}
\newmdtheoremenv{corollary}{Corollary}
\newmdtheoremenv{theorem}{Theorem}
\newmdtheoremenv{lemma}{Lemma}

\title{Non-Exchangeable Conformal Risk Control}

\author{António Farinhas~\textsuperscript{1,2}, Chrysoula Zerva~\textsuperscript{1,2}, Dennis Ulmer~\textsuperscript{3,4}, André F. T. Martins~\textsuperscript{1,2,5}\\
\textsuperscript{1}Instituto de Telecomunicações, \\
\textsuperscript{2}Instituto Superior Técnico, Universidade de  Lisboa (Lisbon ELLIS Unit), \\
\textsuperscript{3}IT University of Copenhagen, 
~ \textsuperscript{4}Pioneer Centre for Artificial Intelligence , 
~ \textsuperscript{5}Unbabel\\
\fontsize{8}{10}\texttt{\{antonio.farinhas,chrysoula.zerva,andre.t.martins\}@tecnico.ulisboa.pt},\\
\fontsize{8}{10}\texttt{dennis.ulmer@mailbox.org}\\
}

\iclrfinalcopy 
\begin{document}

\maketitle

\begin{abstract}
Split conformal prediction has recently sparked great interest due to its ability to provide formally guaranteed uncertainty sets or intervals for predictions made by black-box neural models, ensuring a predefined probability of containing the actual ground truth.
While the original formulation assumes data exchangeability, some extensions handle non-exchangeable data, which is often the case in many real-world scenarios.
In parallel, some progress has been made in conformal methods that provide statistical guarantees for a broader range of objectives, such as bounding the best $F_1$-score or minimizing the false negative rate in expectation.
In this paper, we leverage and extend these two lines of work by proposing \emph{non-exchangeable conformal risk control}, which allows controlling the expected value of any monotone loss function when the data is not exchangeable. 
Our framework is flexible, makes very few assumptions, and allows weighting the data based on its
\rebuttal{relevance for a given test example;}
a careful choice of weights may result on tighter bounds, making our framework useful in the presence of change points, time series, or other forms of distribution drift. 
Experiments with both synthetic and real world data show the usefulness of our method.
\end{abstract}

\section{Introduction}

As the use of machine learning systems for automated decision-making becomes more widespread, the demand for these systems to produce reliable and trustworthy predictions has grown significantly.
In this context, conformal prediction \citep{papadopoulos2002inductive, vovk2005algorithmic} has recently resurfaced as an attractive framework.
Instead of providing a single output, this framework creates prediction sets or intervals that inherently account for uncertainty.
These sets come with a statistical guarantee known as \textbf{coverage}, which ensures that they contain the ground truth in expectation, thereby providing a formal promise of reliability.

The standard formulation of conformal prediction has, however, important limitations. First, it assumes that all data is \textbf{exchangeable}, a condition which is often violated in practice (\textit{e.g.}, when there is correlation over time or space). Second, while the predicted sets/intervals provide guarantees on coverage, they do not bound arbitrary losses, some of which may be more relevant for the situation at hand (\textit{e.g.}, the $F_1$-score or the false negative rate in multilabel classification problems).
Several works have been proposed to improve over these two shortcomings, namely through non-exchangeable conformal prediction \citep{tibshirani2019covariate, gibbs2021adaptive, barber2022conformal} and conformal risk control \citep[CRC]{bates2021distribution, angelopoulos2022conformal}.
In this paper, we extend these lines of research and propose \textbf{non-exchangeable conformal risk control} (non-X CRC).
Our main contributions are:
\begin{itemize}[leftmargin=0cm,labelindent=0cm,itemindent=.5cm,topsep=0.5pt,itemsep=.0ex,partopsep=.5ex,parsep=1ex]
    \item We propose a new method for conformal risk control that provides formal guarantees when the data is not exchangeable,  while also achieving the same guarantees as existing methods if the data is in fact exchangeable (see \cref{tab:comparison-methods} where we position our work in the literature);
    \item \cref{theorem:nonx-crc} establishes a new bound on the expected loss (assumed to be monotonic and bounded), allowing weighting the calibration data based on its
    \rebuttal{relevance for a given test example;}
    \item We demonstrate the usefulness of our framework on three tasks: multilabel classification on synthetic data by minimizing the false negative rate; monitoring electricity usage by minimizing \rebuttal{the} $\lambda$-insensitive absolute loss; and open-domain question answering by bounding the best $F_1$-score.\footnote{Our code is available at \url{https://github.com/deep-spin/non-exchangeable-crc}.} 
\end{itemize}

Throughout the paper, we use the following definition of exchangeable data distribution, which is a weaker assumption than independent and identically distributed (i.i.d.) data.

\begin{definition}[\textbf{Exchangeable data distribution}]
    Let $\mathcal{X}$ and $\mathcal{Y}$ designate input and output spaces. 
    A data distribution in $\mathcal{X} \times \mathcal{Y}$ is said to be exchangeable if and only if we have $\mathbb{P}((X_{\pi(1)},Y_{\pi(1)}),\ldots,(X_{\pi(n)},Y_{\pi(n)})) = \mathbb{P}((X_1,Y_1),\ldots,(X_n,Y_n))$ for any finite sample $\{(X_i, Y_i)\}_{i=1}^n \subseteq \mathcal{X} \times \mathcal{Y}$ and any permutation function $\pi$.
    Note that if the data distribution is i.i.d., then it is also exchangeable, since $\mathbb{P}((X_1,Y_1),\ldots,(X_n,Y_n)) = \prod_{i=1}^n\mathbb{P}((X_i, Y_i))$.
\end{definition}

\begin{table*}[t]
\caption{Our framework combines two approaches,  non-exchangeable conformal prediction and conformal risk control. Through this combination we are able to control the expected value of arbitrary monotonic loss functions when the data is not exchangeable, extending both frameworks.
}
\label{tab:comparison-methods}
\begin{small}
\begin{center}
\begin{tabular}{lll}
\toprule
Method & Data assumptions & Loss\\
\midrule
\citet{papadopoulos2002inductive} &exchangeable \textcolor{white}{\XSolidBrush} &miscoverage\\
\citet{barber2022conformal} &\XSolidBrush &miscoverage \\
\citet{angelopoulos2022conformal} &exchangeable \textcolor{white}{\XSolidBrush} &nonincreasing, arbitrary  \\
\citet[Prop.~3]{angelopoulos2022conformal} & covariate shift, known likelihood ratio 
\textcolor{white}{\XSolidBrush} &nonincreasing, arbitrary \\
\textbf{This paper} &\XSolidBrush &nonincreasing, arbitrary \\
\bottomrule
\end{tabular}
\end{center}
\end{small}
\end{table*}

\section{Background}
\label{sec:Background}

We start by providing background on conformal prediction \citep{papadopoulos2002inductive, vovk2005algorithmic} in \cref{sec:conformal-prediction}. We then discuss recent extensions of the framework---\cref{sec:non-exchangeable-conformal-prediction} discusses the case where the data is non-exchangeable \citep{barber2022conformal}, which is often the case when models are deployed in practice.
Another extension pivots from guaranteeing coverage to instead constraining the expected value of any monotone loss function \citep{angelopoulos2022conformal}, useful for tasks in which the natural notion of error is not miscoverage (\cref{sec:conformal-risk-control}).

\subsection{Conformal prediction}\label{sec:conformal-prediction}

Although other methods exist, this paper focuses on \emph{split} conformal prediction (\citealp{papadopoulos2002inductive}; hereinafter referred to simply as \emph{conformal prediction}).
We start with a pretrained model and measure its performance on a calibration set $\{(X_i,Y_i)\}_{i=1}^{n}$ of paired examples.
Under the assumption of exchangeable data $\{(X_i,Y_i)\}_{i=1}^{n+1}$, conformal prediction constructs prediction sets with the following coverage guarantee:

\begin{equation}\label{eq:conformal-prediction-guarantees}
    \mathbb{P}\big(Y_{n+1}\in \mathcal{C}(X_{n+1})\big) \ge 1 - \alpha,
\end{equation}

where $(X_{n+1},Y_{n+1})$ is a new data point and $\alpha$ a predefined confidence level.
This is accomplished through the following steps:
Let $s(x,y) \in \mathbb{R}$ be a non-conformity score function, where larger scores indicate worse agreement between $x$ and $y$. We compute the value $\hat{q}$ as the $\sfrac{1}{n}\lceil (n+1)(1-\alpha)\rceil$ quantile of the calibration scores and construct a prediction set as follows:

\begin{equation}
\label{eq:prediction-set-cp}
    \mathcal{C}\big(X_{n+1}\big) = \big\{y:\ s(X_{n+1}, y) \le \hat{q}\big\}.
\end{equation}

This prediction set satisfies the coverage guarantee in \cref{eq:conformal-prediction-guarantees}, see \textit{e.g.}, \citealp[App. D]{angelopoulos2021gentle} for a proof.
While this guarantee helps to ensure a certain reliability of the calibrated model, the assumption of exchangeable data is often not true when models are deployed in practice, \textit{e.g.}, due to distribution drift in time series or correlations between different data points.

\subsection{Non-exchangeable conformal prediction}\label{sec:non-exchangeable-conformal-prediction}

\rebuttal{Let us now consider 
prespecified weights $\{w_i\}_{i=1}^n \in [0, 1]^n$} and define $\tilde{w}_i := w_i / (1 + \sum_{i=1}^N w_i)$.
We take a look at a generalization of conformal prediction put together by \mcite{barber2022conformal}, which provides the following coverage guarantee, also valid when exchangeability is violated:
\begin{equation}\label{eq:non-exchangeable-conformal-prediction-guarantee}
    \mathbb{P}\big(Y_{n+1}\in \mathcal{C}(X_{n+1})\big) \ge 1 - \alpha - \sum_{i=1}^n \tilde{w}_id_{\mathrm{TV}}(Z, Z^i),
\end{equation}
where $Z := (X_1, Y_1), \ldots, (X_n, Y_n), (X_{n+1}, Y_{n+1})$ is a sequence of $n$ calibration examples followed by a test example, $Z^i$ denotes $Z$ after swapping \rebuttal{$(X_i, Y_i)$ with $(X_{n+1}, Y_{n+1})$}, and $d_{\mathrm{TV}}(Z, Z^i)$ is the total variation (TV) distance between $Z$ and $Z^i$. 
This is accomplished by using
\begin{equation}\label{eq:non-exchangeable-quantile}
    \hat{q} = \inf \Big\{q: \sum_{i=1}^N \tilde{w}_i \mathbf{1}\big\{s_i\le q \big\} \ge 1 - \alpha \Big\}
\end{equation}
to construct prediction sets the same way as in \cref{eq:prediction-set-cp}. See \citet[\S4]{barber2022conformal} for a proof.
It is worth noting that this method recovers standard conformal prediction when $\{w_i\}_{i=1}^n = 1$. Besides, if the data is exchangeable, then the distribution of $Z$ is equal to the distribution of $Z^i$, and thus using a weighted procedure does not hurt coverage according to \cref{eq:non-exchangeable-conformal-prediction-guarantee}, since $d_{\mathrm{TV}}(Z, Z^i)=0$ for all $i$. 
Intuitively, the ``closer'' to exchangeable the data is, the smaller the last term will be in \cref{eq:non-exchangeable-conformal-prediction-guarantee}. By choosing wisely the weights $w_i$---e.g., by setting large weights to calibration points $(x_i, y_i)$ such that $Z$ and $Z^i$ are similarly distributed and smaller weights otherwise---tighter bounds can be obtained. For example, in time series data we may want to place larger weights on more recent observations.

\subsection{Conformal risk control}\label{sec:conformal-risk-control}

Let us now consider an additional parameter $\lambda$ and construct prediction sets of the form $\mathcal{C}_\lambda(\cdot)$, where larger $\lambda$ yield larger prediction sets, \textit{i.e.}, \rebuttal{$\lambda \le \lambda' \,\,\Longrightarrow\,\, \mathcal{C}_\lambda(.) \subseteq  \mathcal{C}_{\lambda'}(.)$} (see \citet[\S4.3]{angelopoulos2021gentle} for an example).
\rebuttal{Let $\ell$ be an arbitrary (bounded) loss function that shrinks as $\mathcal{C}(X_{n+1})$ grows (\textit{i.e.}, that is \textbf{monotonically nonincreasing} with respect to $\lambda$).}
We switch from conformal methods that provide prediction sets that bound the miscoverage $\mathbb{P}\big(Y_{n+1}\notin \mathcal{C}(X_{n+1})\big) \le \alpha$ to conformal risk control \citep{angelopoulos2022conformal}, which provides guarantees of the form
\begin{equation}\label{eq:conformal-risk-control-guarantee}
    \mathbb{E}\Big[\underbrace{\ell(\mathcal{C}(X_{n+1}), Y_{n+1})}_{L_{n+1}(\hat{\lambda})}\Big] \le \alpha.
\end{equation}
This is accomplished as follows. Let $L_i(\lambda) = \ell(\mathcal{C}_\lambda(X_i), Y_i), ~i=1,\ldots,n+1$, with $L_i: \Lambda \rightarrow (-\infty, B] \,$ and $\lambda_{\text{max}} := \sup \Lambda$, be an exchangeable collection of nonincreasing functions of $\lambda$.
Choosing an optimal $\hat{\lambda}$ as 
\begin{equation}
    \label{eq:crc-lhat}
    \hat{\lambda} = \inf \Big\{\lambda:\ \frac{n}{n + 1} \hat{R}_n(\lambda) + \frac{B}{n+1} \le \alpha \Big\}, \quad
    \hat{R}_n(\lambda) = \frac{1}{n}\sum_{i=1}^n L_i(\lambda),
\end{equation}
yields the guarantee in \cref{eq:conformal-risk-control-guarantee}, see \citet[\S2]{angelopoulos2022conformal} for a proof.
When $\ell(\mathcal{C}(X_{n+1}), Y_{n+1}) = \mathbf{1}\big\{Y_{n+1}\notin \mathcal{C}(X_{n+1})\big\}$ is the miscoverage loss, we recover standard conformal prediction (\cref{sec:conformal-prediction}). 
Note that, as required, this loss is nonincreasing. Other nonincreasing losses include the false negative rate,  
$\lambda$-insensitive absolute error, and the best token-level $F_1$-loss, all of which used in our experiments in \S\ref{sec:Experiments}. 
A limitation of the construction presented in this section is that it relies on the assumption of data exchangeability, which might be violated in practical settings. Our work circumvents this requirement, as we show next.

\section{Non-exchangeable conformal risk control}\label{sec:method}

Up to this point, we have described how to construct prediction sets/intervals with coverage guarantees for non-exchangeable data, in \cref{sec:non-exchangeable-conformal-prediction}, and how to control the expected value of arbitrary monotone loss functions, when the data is exchangeable, in \cref{sec:conformal-risk-control}.
Using the same notation as before, we now present our method, \emph{non-exchangeable conformal risk control}, which puts together these parallel lines of research, providing guarantees of the form:
\begin{equation}\label{eq:non-exchangeable-conformal-risk-control-guarantee}
    \mathbb{E}[L(\hat{\lambda}; (X_{n+1}, Y_{n+1}))] \le \alpha + (B-A)\sum_{i=1}^n \tilde{w}_i d_{\mathrm{TV}}(Z, Z^i),
\end{equation}
where we additionally assume $A<B<\infty$ to be a lower bound on $L_i: \Lambda \rightarrow [A, B]$. Let us define $N_w := \sum_{i=1}^N w_i$. \cref{eq:non-exchangeable-conformal-risk-control-guarantee} is obtained by choosing an optimal $\hat{\lambda}$ as
\begin{equation}
    \hat{\lambda}  = \mathrm{inf} \left\{ \lambda : \frac{N_w}{N_w+1} \hat{R}_n(\lambda) + \frac{B}{N_w+1} \le \alpha \right\}, \quad \hat{R}_n(\lambda) = \frac{1}{N_w}\sum_{i=1}^n w_i L(\lambda; (x_i, y_i)).
\end{equation}

We can see how \cref{eq:non-exchangeable-conformal-risk-control-guarantee} simultaneously mirrors both \cref{eq:non-exchangeable-conformal-prediction-guarantee} and \cref{eq:conformal-risk-control-guarantee}: for an optimal choice of $\lambda$, the expected risk for a new test point is bounded by $\alpha$ plus an extra loosening term that depends on the normalized weights $\{w_i\}_{i=1}^n$ and on the total variation distance between \rebuttal{$Z$ and $Z^i$}.
When the data is in fact exchangeable, we have again $d_{\mathrm{TV}}(Z, Z^i)=0$ for all $i$, and we recover \cref{eq:conformal-risk-control-guarantee}, \textit{i.e.}, our method achieves the same coverage guarantees as standard conformal risk control.
\rebuttal{Although our theoretical bound in \cref{eq:non-exchangeable-conformal-risk-control-guarantee} holds for any choice of weights, 
this result is only useful
when the loosening term
is small, \textit{i.e.}, if we
choose small weights $w_i$ for data points $Z^i$ with large total variation distance $d_\mathrm{TV}(Z, Z^i)$. While the true value of this term is typically unknown, in some situations, such as distribution drift in time series, we expect it to decrease with $i$, motivating the choice of weights that increase with $i$. The same principle can be applied in other domains (\textit{e.g.}, for spatial data, one may place higher weights to points close in space to the test point). We come back to this point in \cref{subsec:choose-weights}.
}

\rebuttal{
The result in \cref{eq:non-exchangeable-conformal-risk-control-guarantee} is valid when the weights are fixed, \textit{i.e.}, data-independent.
However, our result still applies in the case of data-dependent weights $w_i=w(X_i, X_{n+1})$
if we replace $\sum_{i=1}^n \tilde{w}_i d_{\mathrm{TV}}(Z, Z^i)$ by $\mathbb{E}\left[\sum_{i=1}^n \tilde{w}_i d_{\mathrm{TV}}(Z, Z^i|w_1,\ldots,w_n)\right]$ (see \citet[\S4.5]{barber2022conformal} for more information).
We experiment with this approach in \cref{subsec:exp-qa}, where $w_i$ is a function of the embedding similarity between $X_i$ and $X_{n+1}$, showing that the new bound is still useful in practice.}

\subsection{Formal guarantees}\label{eq:formal-guarantees}

Now that we have presented an overview of 
our method, we proceed to providing a formal proof for the guarantee in \cref{{eq:non-exchangeable-conformal-risk-control-guarantee}}.
We begin with a lemma, proved in \cref{sec:proof_lemma}, that establishes a TV bound that extends the one introduced by \citet{barber2022conformal}:

\begin{lemma}\label{lemma:tv_bound}
    Let $f: S \rightarrow [A,B] \subset \mathbb{R}$ be a bounded function on a measurable space $(S, \mathcal{A})$ (where $\mathcal{A} \subseteq 2^S$ is a $\sigma$-algebra) and let $P$ and $Q$ be two probability measures on $(S, \mathcal{A})$. 
    Then
    \begin{align}
        |\mathbb{E}_P[f] - \mathbb{E}_Q[f]| \le (B-A) d_{\mathrm{TV}}(P, Q).
    \end{align}
\end{lemma}

Note that when $f(t) = \mathbf{1}\big\{t \in V\big\}$ for some event $V \in \mathcal{A}$, the left-hand side becomes $|P(V) - Q(V)|$ and we recover the bound used in the proof of \citet[\S{6.2}]{barber2022conformal}.

We now state the main result. The proof technique is similar to that of \citet{barber2022conformal}, but instead of modeling the event of a variable belonging to a ``strange set'', we model expectations of loss functions that depend on a calibration variable. See \cref{sec:proof_theorem} for the full proof.
\begin{theorem}[\textbf{Non-exchangeable conformal risk control}]
\label{theorem:nonx-crc}
Assume that for all $(x,y) \in \mathcal{X}\times \mathcal{Y}$ the loss $L(\lambda; (x,y))$ is nonincreasing in $\lambda$ and bounded as $A \le L(\lambda; (x,y)) \le B$ for any $\lambda$. 
Let $$Z := (X_1, Y_1), \ldots, (X_n, Y_n), (X_{n+1}, Y_{n+1})$$ be a sequence of $n$ calibration examples followed by a test example, and let 
$w_1, \ldots, w_n \in [0,1]^n$ be \rebuttal{\textbf{data-independent}} weights. 
Define $N_w = \sum_{i=1}^n w_i$,  $\tilde{w}_i = w_i / (N_w + 1)$ for $i \in [n]$ and $\tilde{w}_{n+1} = 1 / (N_w + 1)$. 
Let $\alpha \in [A,B]$ be the maximum tolerable risk, and  define 
\begin{align}\label{eq:lambda_hat}
    \hat{\lambda} &= \mathrm{inf} \left\{ \lambda : \frac{N_w}{N_w+1} \hat{R}_n(\lambda) + \frac{B}{N_w+1} \le \alpha \right\},
\end{align}
where $\hat{R}_n(\lambda)$ is the weighted empirical risk in the calibration set:
\begin{align}
    \hat{R}_n(\lambda) = \frac{1}{N_w}\sum_{i=1}^n w_i L(\lambda; (x_i, y_i)).
\end{align}
Then, we have
\begin{align}
    \mathbb{E}[L(\hat{\lambda}; (X_{n+1}, Y_{n+1}))] \le \alpha + (B-A)\sum_{i=1}^n \tilde{w}_i d_{\mathrm{TV}}(Z, Z^i),
\end{align}
where $Z^{i}$ is obtained from $Z$ by swapping $(X_i, Y_i)$ and $(X_{n+1}, Y_{n+1})$. 
\end{theorem}

The next section illustrates how we can make practical use of this result to minimize loss functions beyond the miscoverage loss in the presence of non-exchangeable data distributions.

\rebuttal{
\subsection{How to choose weights}
\label{subsec:choose-weights}

To make practical use of \cref{theorem:nonx-crc}, we need a procedure to choose the weights $w_i$.
We next suggest a strategy based on regularized minimization of the coverage gap $g(\tilde{w}_1, ..., \tilde{w}_n) := (B-A)\sum_{i=1}^n \tilde{w}_i d_\mathrm{TV}(Z, Z^i)$ via the maximum entropy principle \citep{jaynes1957information}. 
Note first that simply minimizing this gap would lead to $\tilde{w}_i=0$ for all $i \in [n]$ and $\tilde{w}_{n+1}=1$, which ignores all the calibration data and leads to an infeasible $\hat{\lambda}$ in \cref{eq:crc-lhat}. 
In general, if all weights ${w}_i$ are too small, this leads to a very large $w_{n+1}$ and an unreasonably large $\hat{\lambda}$. 
On the other extreme, having all weights too large (e.g. $w_i=1$ for all $i$, which leads to $\tilde{w}_i = 1/(n+1)$ for $i \in [n+1]$) ignores the non-exchangeability of the data and may lead to a large coverage gap. 
Therefore, it is necessary to find a good balance between ensuring a small coverage gap but at the same time ensuring that the distribution $\tilde{w}_1, ..., \tilde{w}_{n+1}$ is not too peaked, i.e., that it has sufficiently high entropy. 
Since by definition, we must have $\tilde{w}_{n+1} \ge \tilde{w}_{i}$ for all $i \in [n]$, this can be formalized as the following regularized minimization problem:
\begin{align}\label{eq:maxent}
    & \min_{\tilde{w}_{1}, ..., \tilde{w}_{n+1}} (B-A)\sum_{i=1}^n \tilde{w}_i d_\mathrm{TV}(Z, Z^i) - \beta H(\tilde{w}_{1}, ..., \tilde{w}_{n+1}) \nonumber\\
    & \text{subject to $\sum_{i=1}^{n+1} \tilde{w}_i = 1$ and  $0 \le \tilde{w}_i \le \tilde{w}_{n+1}$ for all $i \in [n]$,}
\end{align}
where $H(\tilde{w}_{1}, ..., \tilde{w}_{n+1}) = -\sum_{i=1}^{n+1} \tilde{w}_{i}\log \tilde{w}_{i}$ is the entropy function and $\beta > 0$ is a temperature parameter. 
The solution of this problem is  
$\tilde{w}_i \propto \exp(-\beta (B-A)d_\mathrm{TV}(Z, Z^i))$ for $i \in [n+1]$. 

Although in general $d_\mathrm{TV}(Z, Z^i)$ is not known, it is possible in some scenarios to bound or to estimate this quantity: for example, when variables are independent but not identically distributed, it can be shown that $d_\mathrm{TV}(Z, Z^i) \le 2d_\mathrm{TV}(Z_i, Z_{n+1})$ \citep[Lemma 1]{barber2022conformal}; and it is possible to upper bound the total variation distance as a function of the (more tractable and amenable to estimation) Kullback-Leibler divergence, e.g., via Pinsker's or Bretagnolle-Huber's inequalities \citep{bretagnolle1979estimation,csiszar2011information}, which may provide good heuristics. 
For example, in a time series under a distribution shift scenario bounded with a Lipschitz-type condition  $d_\mathrm{TV}(Z_i,Z_{n+1}) \le \epsilon(n+1-i)$ for some $\epsilon>0$ (see e.g. \citep[\S 4.4]{barber2022conformal}), we could replace $d_{\mathrm{TV}}(Z, Z^i)$ in \cref{eq:maxent} by this upper bound 
to obtain the maxent solution $\tilde{w}_i \propto \exp(-\beta \epsilon (n+1-i)) = \rho^{n+1-i}$, where $\rho = \exp(-\beta \epsilon) \in (0,1)$. This exponential decay of the weights was suggested by \citep{barber2022conformal}; our maximum entropy heuristic provides further justification for that choice. We use this strategy in some of our experiments in \S\ref{sec:Experiments}.}

\section{Experiments}
\label{sec:Experiments}

In this section, we turn to demonstrating the validity of our theoretical results in three different tasks using different nonincreasing losses: a \textbf{multilabel classification} problem using synthetic time series data, minimizing the false negative rate (\cref{sec:experiment-synthetic}), a problem involving \textbf{monitoring electricity usage}, minimizing the $\lambda$-insensitive absolute loss (\cref{subsec:exp-electricity}), and an \textbf{open-domain question answering} (QA) task, where we control the best token-level $F_1$-score (\cref{subsec:exp-qa}). 
Throughout, we report our method alongside a conformal risk control (CRC) baseline that predicts $\hat{\lambda}$ following \cref{eq:crc-lhat}.

\subsection{Multilabel classification in a time series}
\label{sec:experiment-synthetic}

We start by validating our approach on 
synthetic data, before moving to real-world data in the following subsections. 
To this end, we modified the synthetic regression experiment of \citet[\S5.1]{barber2022conformal} to turn it into a multilabel classification problem with up to $M=10$ different labels.
We consider three different setups:
\begin{enumerate}
    \item \textbf{Exchangeable (i.i.d.) data:} We sample $N=2000$ i.i.d. data points $(X_i,Y_i) \in \mathbb{R}^{M} \times \mathbb{R}^{M}$. We sample $X_i$ from a Gaussian distribution, $X_i \mathop{\sim}\limits^{\mathrm{iid}} \mathcal{N} (\mathbf{0},\bm{I}_{M})$, and we set $Y_i \sim \mathbf{sign}(\bm{W} X_i + \bm{b} + .1\mathcal{N}(\mathbf{0},\bm{I}_{M}))$. The coefficient matrix 
    $\bm{W}$ is set to the identity matrix $\bm{I}_{M}$ and the biases to $\bm{b} = -\mathbf{0.5}$, to encourage a sparse set of labels.
    \item \textbf{Changepoints:} We follow
    setting (1) and sample $N=2000$ i.i.d. data points $(X_i,Y_i)$, setting $X_i \mathop{\sim}\limits^{\mathrm{iid}} \mathcal{N} (\mathbf{0},\bm{I}_{M}) $ and $Y_i \sim \mathbf{sign}(\bm{W}^{(k)} X_i + \bm{b} + .1\mathcal{N}(\mathbf{0}, \bm{I}_{M}))$, again with $\bm{b}=-\mathbf{0.5}$. We start with the same coefficients $\bm{W}^{(0)}=\bm{I}_{M}$ and for every changepoint $k>0$ we rotate the coefficients such that $\bm{W}^{(k)}_{i,j} = \bm{W}^{(k-1)}_{i-1,j}$ for $i > 1$ and $\bm{W}^{(k)}_{1,j} = \bm{W}^{(k-1)}_{M,j}$. 
    Following \citet{barber2022conformal}, we use two changepoints ($k=2$) at timesteps $500$ and $1500$.
    \item \textbf{Distribution drift:} We follow setting (2) and sample $N=2000$ i.i.d. data points $(X_i,Y_i)$, with $X_i \mathop{\sim}\limits^{\mathrm{iid}} \mathcal{N} (\mathbf{0},\bm{I}_{M}) $ and $Y_i \sim \mathbf{sign}(\bm{W}^{(k)} X_i + \bm{b} + .1\mathcal{N}(\mathbf{0},\bm{I}_M))$, with $\bm{b}$ as above. Again, we start with $\bm{W}^{(0)}=\bm{I}_{M}$ but now we set
    $\bm{W}^{(N)}$ to the last  matrix of setting (2). We then compute each intermediate $\bm{W}^{(k)}$ by linearly interpolating between $\bm{W}^{(0)}$ and $\bm{W}^{(N)}$.
\end{enumerate}

After a warmup period of $200$ time points, at each time step $n=200,\ldots,N-1$ we assign odd indices to the training set, even indices to the calibration set, and we let $X_{n+1}$ be the test point. 
We fit $M$ independent logistic regression models to the training data to obtain predictors for each label; we let $f_m(X_i)$ denote the estimated probability of the $m\textsuperscript{th}$ label according to the model. Based on this predictor, we define prediction sets 
$\mathcal{C}_\lambda(X_i) := \{m \in [M]\,:\, f_m(X_i) \ge 1-\lambda\}$.
We compare standard CRC with non-exchangeable (non-X) CRC, for which we
use weights $w_i=0.99^{n+1-i}$ and predict $\hat{\lambda}$ following \cref{eq:lambda_hat}.
In both cases, we minimize the \textbf{false negative rate} (FNR):\footnote{With some abuse of notation, we use $Y_i \subseteq \{1, \ldots, M\}$ to denote the set of gold labels with value $+1$.}
\begin{equation}
    L(\lambda;(X_i,Y_i)) = 1 - \frac{|Y_i \cap \mathcal{C}_{\lambda}(X_i)|}{|Y_i|}. 
\end{equation}
Note that this loss is nonincreasing in $\lambda$, as required. 
\cref{app:experiment-synthetic} contains additional experiments considering $\lambda$ to be the number of active labels and using $\mathcal{C}_\lambda(X_i) = \text{top-$\lambda$}(\bm{f}(X_i))$.

\begin{figure}[t]
\begin{center}
\includegraphics[width=1.0\columnwidth]{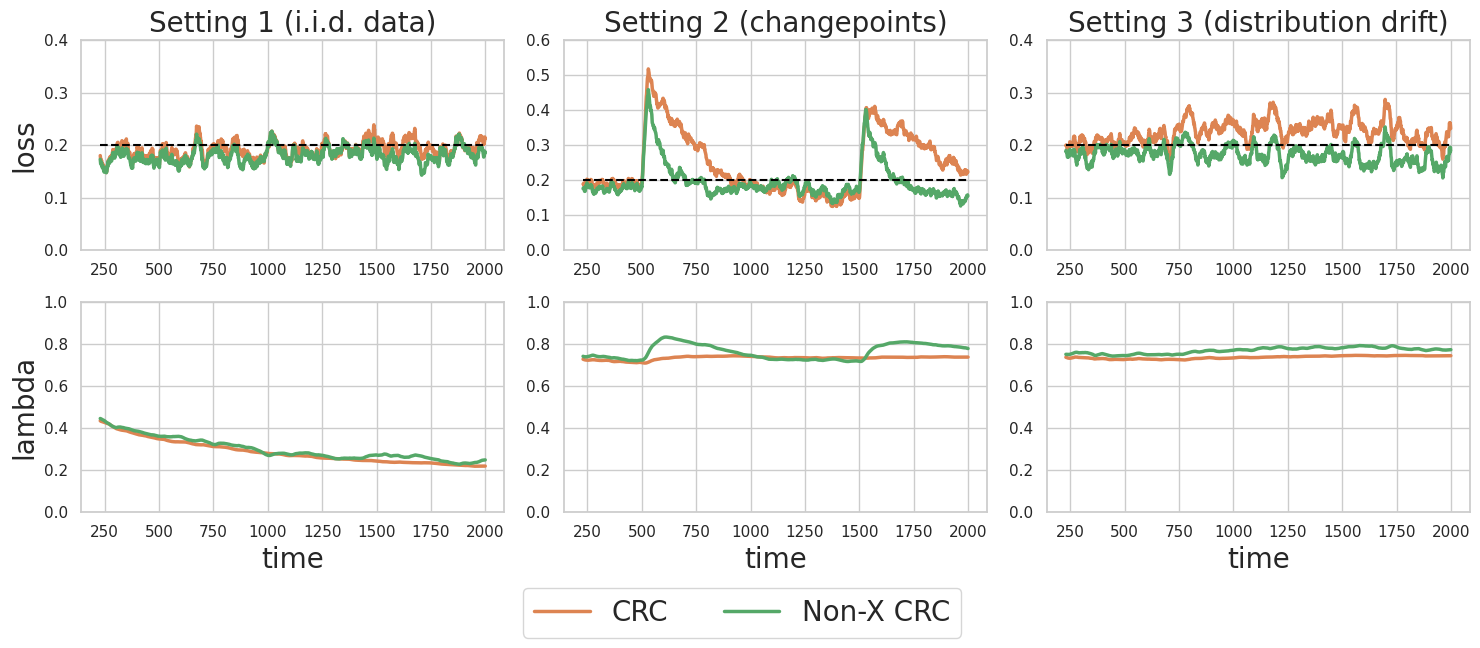}
\caption{Average loss (top) and $\hat{\lambda}$ (bottom) over $10$ independent trials for settings (1), (2), and (3). We smooth all the curves by taking a rolling average with a window of $30$ time points.}
\label{fig:plot-simulations}
\end{center}
\end{figure}

\cref{fig:plot-simulations} shows results averaged across $10$ independent trials for $\alpha=0.2$\rebuttal{, summarized in \cref{tab:scalar-statistics}.}
We see that the performance of both methods is comparable when the data is i.i.d, with non-X CRC being slightly more conservative. However, when the data is not exchangeable due to the presence of changepoints or distribution drift, our proposed method is considerably better.
In particular, after the changepoints in setting (2), non-X CRC is able to achieve the desired risk level more rapidly; in setting (3), the performance of standard CRC gradually drops over time---a problem that can be mitigated by accounting for non-exchangeability introduced by the distribution drift. \rebuttal{Importantly, while the average risk is above the predefined threshold for standard CRC for settings (2) and (3) ($0.246$ and $0.225$, respectively), our method achieves the desired risk level on average ($0.196$ and $0.182$, respectively).}

\begin{table*}[t]
\caption{\rebuttal{Scalar statistics (mean/median) for settings (1), (2), and (3) for the multilabel classification problem using synthetic time series data reported in \cref{sec:experiment-synthetic}.}}
\label{tab:scalar-statistics}
\begin{small}
\begin{center}
\begin{tabular}{lccc}
\toprule
\rebuttal{Method} & \rebuttal{Setting 1 (i.i.d. data)}  & \rebuttal{Setting 2 (changepoints)} & \rebuttal{Setting 3 (distribution drift)}\\
\midrule
\rebuttal{CRC} &\rebuttal{0.191~/~0.183} & \rebuttal{0.246~/~0.228} &\rebuttal{0.225~/~0.218} \\
\rebuttal{non-X CRC} &\rebuttal{\bf0.181~/~0.175} &\rebuttal{\bf0.196~/~0.183} &\rebuttal{\bf0.182~/~0.175} \\
\bottomrule
\end{tabular}
\end{center}
\end{small}
\end{table*}

\subsection{Monitoring electricity usage}
\label{subsec:exp-electricity}

We use the ELEC2 dataset \citep{Harries1999SPLICE2CE}, which tracks electricity transfer between two states in Australia, considering the subset of the data used by \citet{barber2022conformal}, which contains $3444$ time points. The data points correspond to the 09:00am - 12:00pm timeframe and we use the price (\texttt{nswprice}, \texttt{vicprice}) and demand (\texttt{nswdemand}, \texttt{vicdemand}) variables as input features, $x_i$ to predict the target \texttt{transfer} values $y_i$. We also consider a randomly permuted version of the dataset such that the exchangeability assumption is satisfied.
We use the same definitions and settings of \cref{sec:experiment-synthetic}, but this time we fit a least squares regression model to predict the \texttt{transfer} values, $\hat{y}_i = f(x_i)$, at each time step. For non-X CRC, we use weights $w_i=0.99^{n+1-i}$ and we also experiment with weighted least-squares regression, placing weights $t_i=w_i$ on each data point (non-X CRC + WLS).  For both standard and non-X CRC we control the residual (distance) with respect to the confidence interval $\mathcal{C}_\lambda(x_i) = [f(x_i) - {\lambda}, f(x_i) + {\lambda}]$, where $f(x_i)$ corresponds to the predicted values for \texttt{transfer}. We use the \textbf{$\lambda$-insensitive absolute loss}, a loss function commonly used in support vector regression \citep{scholkopf1998shrinking, vapnik1999nature}:
\begin{equation}
    L({\lambda};(x_i,y_i)) = \begin{cases}
0, &\text{if }|f(x_i) - y_i| \le \lambda,
\\
|f(x_i)-y_i|-{{\lambda}}, &\text{otherwise}. 
\end{cases}
\end{equation}

We experiment using $\lambda \in [0,1]$ with a step of $0.01$. Since we are using the normalized ELEC2 dataset, \texttt{transfer} takes values in $[0,1]$, thus $L({\lambda};(f(x_i),y_i))$ is bounded by $B=1$. By definition $L({\lambda};(f(x_i),y_i))$ is nonincreasing with respect to ${\lambda}$.

\begin{figure}
    \centering
    \includegraphics[width=\textwidth]{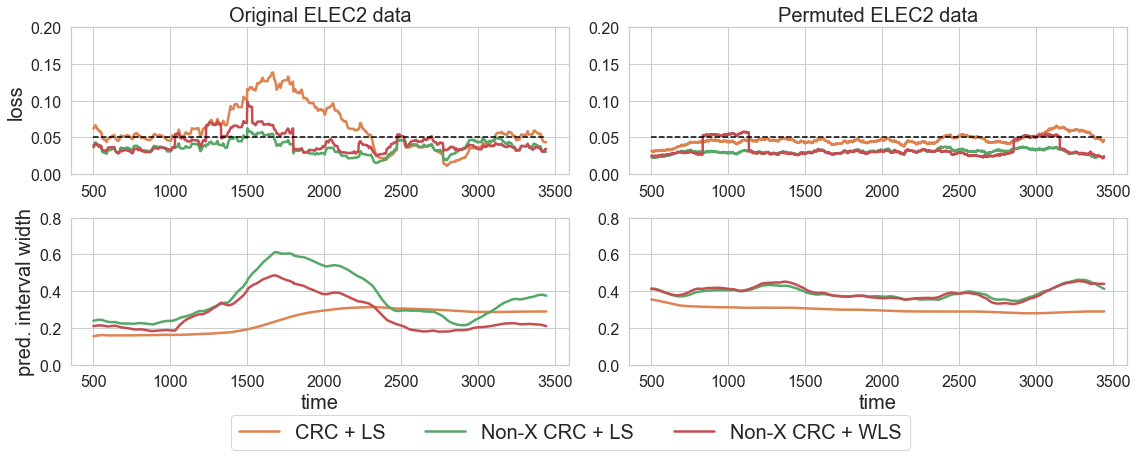}
\caption{Results on ELEC2 data for $\alpha=0.05$ and $\lambda$ defined by the prediction interval width. Presented curves are smoothed by taking a rolling average with a window of 300 data points per timestep.}
\label{fig:wls_elec}
\end{figure}

Fig. \ref{fig:wls_elec} shows results for the aforementioned setup.
We can observe that in the original setting, both non-exchangeable methods approximate well the desired loss threshold even during the timesteps at which the data suffers from distribution drift. Specifically, as observed by \citet{barber2022conformal}, the electricity \emph{transfer} values are more noisy during the middle of the time range and we can see that the standard CRC + LS method underestimates the $\hat{\lambda}$ for these data points resulting in increased loss, above the desired one. With respect to the CRC + WLS setup, we can see 
that it manages to reach the desired loss with a smaller interval width on average, indicating that fitting the weighted least-squares model performs better when the data distribution changes, allowing for smaller $\lambda$ during calibration. 
For the permuted data that simulates the exchangeable data scenario, we can see that all methods perform similarly, reaching the desired loss, as expected.

\subsection{Open-domain question answering}
\label{subsec:exp-qa}

We now shift to open-domain QA, a task that consists in answering factoid questions using a large collection of documents. This is done in two stages, following \citet{angelopoulos2022conformal}: \emph{(i)} a retriever model \citep[DPR]{karpukhin-etal-2020-dense} selects passages from Wikipedia that might contain the answer to the question, and \emph{(ii)} a reader model examines the retrieved contexts and extract text sub-spans that serve as candidate answers.\footnote{Enumerating all possible answers is intractable, and thus we retrieve the top several hundred candidate answers, extracted from the top 100 passages (which is sufficient to control all risks).}

\begin{figure}[t]
\begin{center}
\includegraphics[width=1.0\columnwidth]{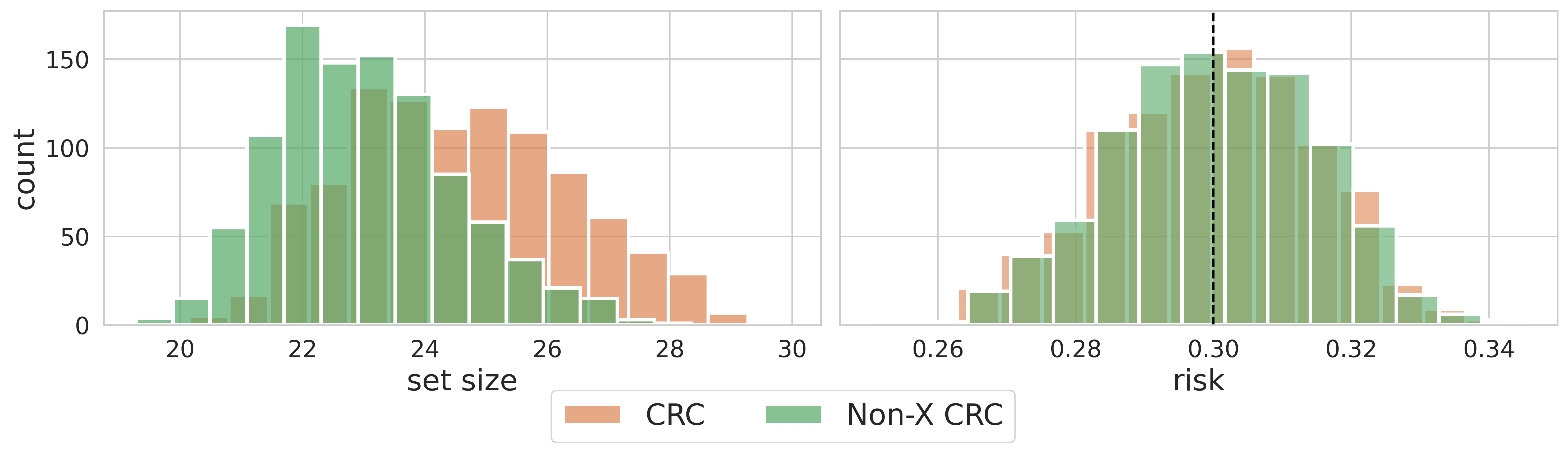}
\caption{$F_1$-score control on the Natural Questions dataset. Average set size (left) and risk (right) over 1000 independent random data splits.\label{fig:plot-qa}}
\end{center}
\end{figure}

Given a vocabulary $\mathcal{V}$, each $X_i \in \mathcal{Z}$ is a question and $Y_i \in \mathcal{Z}^k$ a set of $k$ correct answers, where $\mathcal{Z}:=\mathcal{V}^m$ (we assume that $X_i$ and $Y_i$ are sequences composed of up to $m$ tokens).
We calibrate the \textbf{best token-based $F_1$-score} of the prediction set~\footnote{This is the same loss used by \citet{angelopoulos2022conformal}.}, taken over all pairs of predictions and answers,
\begin{equation}
    L(\lambda; (X_i,Y_i)) = 1- \max \{F_1(a,c): c\in\mathcal{C}_\lambda(X_i), a\in Y_i\},
    \qquad \mathcal{C}_\lambda = \{y: f(X_i, y) \geq \lambda \},
\end{equation}
which is nonincreasing and upper-bounded by $B=1$.
We consider a CRC baseline that predicts $\hat{\lambda}$ following \cref{eq:crc-lhat}.
For non-X CRC, we choose weights $\{w_i\}_{i=1}^{n}$ by computing the dot product between the embedding representations of $\{X_i\}_{i=1}^{n}$ and $X_{n+1}$, obtained using a sentence-transformer model \citep{reimers-gurevych-2019-sentence} designed for semantic search,\footnote{We use the \texttt{multi-qa-mpnet-base-dot-v1} model available at \url{https://huggingface.co/sentence-transformers/multi-qa-mpnet-base-dot-v1}.} and predict $\hat{\lambda}$ following \cref{eq:lambda_hat}. While in standard CRC $\hat{\lambda}$ is the same for each test example, this is not the case for non-X CRC.

\rebuttal{
While \cref{theorem:nonx-crc} requires the 
weights to be
independent of the test example,
we relax this assumption by setting higher weights for questions in a ``neighborhood" of $X_{n+1}$ (see \cref{sec:method}). Intuitively, we could think of a situation where the questions are posed by multiple users, each of which may have a tendency to ask semantically similar questions or from the same domain. In this case, we could choose \textit{a priori} higher weights for closer domains/users without violating this assumption.}

We use the Natural Questions dataset \citep{kwiatkowski-etal-2019-natural, karpukhin-etal-2020-dense},
considering $n=2500$ points for calibration and 
$1110$ for evaluation.
Following \citet{angelopoulos2022conformal}, we use $\alpha=0.3$ and report results over 1000 trials in Fig.~\ref{fig:plot-qa}. While the test risk is similar in both cases ($0.30 \pm 0.015$), the prediction sets of our method are considerably smaller than those of standard CRC ($23.0 \pm 1.47$ vs. $24.6 \pm 1.83$, respectively).
By choosing appropriate weights we can better estimate the set size needed to obtain the desired risk level, while standard CRC tends to overestimate the set size to reach the same value. We thus obtain better estimates of confidence over the predictions.

\section{Related work}

Conformal prediction
\citep{gammerman1998learning, vovk1999machine, saunders1999transduction} 
has proven to be a useful tool for obtaining uncertainty sets/intervals for the predictions of machine learning models, having found a variety of extensions and applications over the years. Among these are \emph{split conformal prediction} \citep{papadopoulos2002inductive}, which does not require retraining the predictor and instead 
uses a held-out dataset and \emph{cross-conformal prediction} \citep{vovk2015cross}, which is 
a hybrid
between split conformal prediction and cross-validation.
Some of these methods have recently been applied in tasks such as language modeling \citep{schuster2022confident}, molecular design \citep{fannjiang2022conformal}, pose estimation \citep{yang2023object}, and image denoising \citep{teneggi2023trust}.

In addition to the works discussed in \cref{sec:Background}, several extensions to non-exchangeable data have been proposed for 
time series \citep{chernozhukov2018exact, chernozhukov2021exact, xu2021conformal, stankeviciute2021conformal, lin2022conformal, zaffran2022adaptive, sun2022copula, schlembach2022conformal, angelopoulos2023conformalpid}, covariate shift \citep{tibshirani2019covariate}, label shift \citep{podkopaev2021label}, and others \citep{cauchois2020robust, gibbs2021adaptive, chernozhukov2021distributional, gibbs2022conformal, oliveira2022split, guan2023localized}.
Moreover, there is recent work aiming at controlling arbitrary risks in an online setting \citep{feldman2022achieving}.
The ideas, assumptions, or formal guarantees in these works are different to ours---we refer the reader to the specific papers for further information.

\citet{angelopoulos2022conformal} touch the case of conformal risk control under \emph{covariate shift} (Proposition~3; without providing any empirical validation), explaining how to generalize the work of \citet{tibshirani2019covariate} to any monotone risk under the strong assumption that the distribution of $Y|X$ is the same for both the training and test data and that the likelihood ratio between $X_\text{test}$ and $X_\text{train}$ is known or can be accurately estimated using a large set of test data. This result is orthogonal to ours.
Besides, they quantify how unweighted conformal risk control degrades when there is an arbitrary distribution shift.
Our work is more general and differs in several significant ways: we allow for an arbitrary design of weights, the bounds can be tighter, and the losses are bounded in $[A,B]$, not necessarily in $[0,B]$.
Specifically, their Proposition~4 is a particular case of our main result (choosing $A=0$ and unitary weights), which we use as a baseline in our experiments.

\section{Conclusions}

We have proposed a new method for conformal risk control, which is still valid when the data is not exchangeable (\textit{e.g.}, due to an arbitrary distribution shift) and provides a tighter bound on the expected loss than that of previous work.
Our simulated experiments illustrate how non-exchangeable conformal risk control effectively provides prediction sets satisfying the risk requirements in the presence of non-exchangeable data (in particular, in the presence of change points and distribution drift), without sacrificing performance if the data is in fact exchangeable. Additional experiments with real data validate the usefulness of our approach.

Our work opens up exciting possibilities for research on
risk control in 
challenging settings. For instance, it is an attractive framework for providing guarantees on the predictions of large language models, being of particular interest in tasks involving 
language generation, medical data \citep{jalali2020deep}, or reinforcement learning \citep{wang2023conformal}, where the i.i.d. assumption does not hold.

\section*{Acknowledgments}
We would like to thank Mário Figueiredo, the SARDINE lab team, and the anonymous reviewers for helpful discussions.
This work was built on open-source software; we acknowledge \citet{python, numpy, scipy, nparray, scikit-learn}, and \cite{pytorch}.
This work was supported by EU's Horizon Europe Research and Innovation Actions (UTTER, contract 101070631), by the project DECOLLAGE (ERC-2022-CoG 101088763), by the Portuguese Recovery and Resilience Plan through project C645008882-00000055 (Center for Responsible AI), and by Fundação para a Ciência e Tecnologia through contract UIDB/50008/2020.

\bibliography{iclr2024_conference}

\begin{thebibliography}{50}
\providecommand{\natexlab}[1]{#1}
\providecommand{\url}[1]{\texttt{#1}}
\expandafter\ifx\csname urlstyle\endcsname\relax
  \providecommand{\doi}[1]{doi: #1}\else
  \providecommand{\doi}{doi: \begingroup \urlstyle{rm}\Url}\fi

\bibitem[Angelopoulos \& Bates(2021)Angelopoulos and Bates]{angelopoulos2021gentle}
Anastasios~N Angelopoulos and Stephen Bates.
\newblock A gentle introduction to conformal prediction and distribution-free uncertainty quantification.
\newblock \emph{arXiv preprint arXiv:2107.07511}, 2021.

\bibitem[Angelopoulos et~al.(2023{\natexlab{a}})Angelopoulos, Bates, Fisch, Lei, and Schuster]{angelopoulos2022conformal}
Anastasios~N. Angelopoulos, Stephen Bates, Adam Fisch, Lihua Lei, and Tal Schuster.
\newblock Conformal risk control, 2023{\natexlab{a}}.

\bibitem[Angelopoulos et~al.(2023{\natexlab{b}})Angelopoulos, Candes, and Tibshirani]{angelopoulos2023conformalpid}
Anastasios~N. Angelopoulos, Emmanuel~J. Candes, and Ryan~J. Tibshirani.
\newblock Conformal pid control for time series prediction, 2023{\natexlab{b}}.

\bibitem[Barber et~al.(2023)Barber, Cand{\`e}s, Ramdas, and Tibshirani]{barber2022conformal}
Rina~Foygel Barber, Emmanuel~J. Cand{\`e}s, Aaditya Ramdas, and Ryan~J. Tibshirani.
\newblock {Conformal prediction beyond exchangeability}.
\newblock \emph{The Annals of Statistics}, 51\penalty0 (2):\penalty0 816 -- 845, 2023.
\newblock \doi{10.1214/23-AOS2276}.
\newblock URL \url{https://doi.org/10.1214/23-AOS2276}.

\bibitem[Bates et~al.(2021)Bates, Angelopoulos, Lei, Malik, and Jordan]{bates2021distribution}
Stephen Bates, Anastasios Angelopoulos, Lihua Lei, Jitendra Malik, and Michael Jordan.
\newblock Distribution-free, risk-controlling prediction sets.
\newblock \emph{J. ACM}, 68\penalty0 (6), sep 2021.
\newblock ISSN 0004-5411.
\newblock \doi{10.1145/3478535}.
\newblock URL \url{https://doi.org/10.1145/3478535}.

\bibitem[Bretagnolle \& Huber(1979)Bretagnolle and Huber]{bretagnolle1979estimation}
Jean Bretagnolle and Catherine Huber.
\newblock Estimation des densit{\'e}s: risque minimax.
\newblock \emph{Zeitschrift f{\"u}r Wahrscheinlichkeitstheorie und verwandte Gebiete}, 47:\penalty0 119--137, 1979.

\bibitem[Cauchois et~al.(2020)Cauchois, Gupta, Ali, and Duchi]{cauchois2020robust}
Maxime Cauchois, Suyash Gupta, Alnur Ali, and John~C Duchi.
\newblock Robust validation: Confident predictions even when distributions shift.
\newblock \emph{arXiv preprint arXiv:2008.04267}, 2020.

\bibitem[Chernozhukov et~al.(2018)Chernozhukov, W\"{u}thrich, and Yinchu]{chernozhukov2018exact}
Victor Chernozhukov, Kaspar W\"{u}thrich, and Zhu Yinchu.
\newblock Exact and robust conformal inference methods for predictive machine learning with dependent data.
\newblock In Sébastien Bubeck, Vianney Perchet, and Philippe Rigollet (eds.), \emph{Proceedings of the 31st Conference On Learning Theory}, volume~75 of \emph{Proceedings of Machine Learning Research}, pp.\  732--749. PMLR, 06--09 Jul 2018.
\newblock URL \url{https://proceedings.mlr.press/v75/chernozhukov18a.html}.

\bibitem[Chernozhukov et~al.(2021{\natexlab{a}})Chernozhukov, W{\"u}thrich, and Zhu]{chernozhukov2021distributional}
Victor Chernozhukov, Kaspar W{\"u}thrich, and Yinchu Zhu.
\newblock Distributional conformal prediction.
\newblock \emph{Proceedings of the National Academy of Sciences}, 118\penalty0 (48):\penalty0 e2107794118, 2021{\natexlab{a}}.
\newblock \doi{10.1073/pnas.2107794118}.
\newblock URL \url{https://www.pnas.org/doi/abs/10.1073/pnas.2107794118}.

\bibitem[Chernozhukov et~al.(2021{\natexlab{b}})Chernozhukov, Wüthrich, and Zhu]{chernozhukov2021exact}
Victor Chernozhukov, Kaspar Wüthrich, and Yinchu Zhu.
\newblock An exact and robust conformal inference method for counterfactual and synthetic controls.
\newblock \emph{Journal of the American Statistical Association}, 116\penalty0 (536):\penalty0 1849–1864, Jun 2021{\natexlab{b}}.
\newblock ISSN 1537-274X.
\newblock \doi{10.1080/01621459.2021.1920957}.
\newblock URL \url{http://dx.doi.org/10.1080/01621459.2021.1920957}.

\bibitem[Csisz{\'a}r \& K{\"o}rner(2011)Csisz{\'a}r and K{\"o}rner]{csiszar2011information}
Imre Csisz{\'a}r and J{\'a}nos K{\"o}rner.
\newblock \emph{Information theory: coding theorems for discrete memoryless systems}.
\newblock Cambridge University Press, 2011.

\bibitem[Fannjiang et~al.(2022)Fannjiang, Bates, Angelopoulos, Listgarten, and Jordan]{fannjiang2022conformal}
Clara Fannjiang, Stephen Bates, Anastasios~N. Angelopoulos, Jennifer Listgarten, and Michael~I. Jordan.
\newblock Conformal prediction under feedback covariate shift for biomolecular design.
\newblock \emph{Proceedings of the National Academy of Sciences}, 119\penalty0 (43):\penalty0 e2204569119, 2022.
\newblock \doi{10.1073/pnas.2204569119}.
\newblock URL \url{https://www.pnas.org/doi/abs/10.1073/pnas.2204569119}.

\bibitem[Feldman et~al.(2022)Feldman, Ringel, Bates, and Romano]{feldman2022achieving}
Shai Feldman, Liran Ringel, Stephen Bates, and Yaniv Romano.
\newblock Achieving risk control in online learning settings, 2022.

\bibitem[Gammerman et~al.(1998)Gammerman, Vovk, and Vapnik]{gammerman1998learning}
Alexander Gammerman, Volodya Vovk, and Vladimir Vapnik.
\newblock Learning by transduction.
\newblock In Gregory~F. Cooper and Seraf{\'{\i}}n Moral (eds.), \emph{{UAI} '98: Proceedings of the Fourteenth Conference on Uncertainty in Artificial Intelligence, University of Wisconsin Business School, Madison, Wisconsin, USA, July 24-26, 1998}, pp.\  148--155. Morgan Kaufmann, 1998.

\bibitem[Gibbs \& Candes(2021)Gibbs and Candes]{gibbs2021adaptive}
Isaac Gibbs and Emmanuel Candes.
\newblock Adaptive conformal inference under distribution shift.
\newblock In M.~Ranzato, A.~Beygelzimer, Y.~Dauphin, P.S. Liang, and J.~Wortman Vaughan (eds.), \emph{Advances in Neural Information Processing Systems}, volume~34, pp.\  1660--1672. Curran Associates, Inc., 2021.
\newblock URL \url{https://proceedings.neurips.cc/paper_files/paper/2021/file/0d441de75945e5acbc865406fc9a2559-Paper.pdf}.

\bibitem[Gibbs \& Candès(2022)Gibbs and Candès]{gibbs2022conformal}
Isaac Gibbs and Emmanuel Candès.
\newblock Conformal inference for online prediction with arbitrary distribution shifts, 2022.

\bibitem[Guan(2022)]{guan2023localized}
Leying Guan.
\newblock Localized conformal prediction: a generalized inference framework for conformal prediction.
\newblock \emph{Biometrika}, 110\penalty0 (1):\penalty0 33–50, Jul 2022.
\newblock ISSN 1464-3510.
\newblock \doi{10.1093/biomet/asac040}.
\newblock URL \url{http://dx.doi.org/10.1093/biomet/asac040}.

\bibitem[Harries(1999)]{Harries1999SPLICE2CE}
Michael Harries.
\newblock Splice-2 comparative evaluation: Electricity pricing.
\newblock In \emph{Technical report, University of New South Wales}, 1999.

\bibitem[Jalali et~al.(2020)Jalali, Lonsdale, Do, Peck, Gupta, Kutty, Ghazarian, Jacobs, Rehman, and Ahumada]{jalali2020deep}
Ali Jalali, Hannah Lonsdale, Nhue Do, Jacquelin Peck, Monesha Gupta, Shelby Kutty, Sharon~R. Ghazarian, Jeffrey~P. Jacobs, Mohamed Rehman, and Luis~M. Ahumada.
\newblock Deep learning for improved risk prediction in surgical outcomes.
\newblock \emph{Scientific Reports}, 10\penalty0 (1):\penalty0 9289, 2020.
\newblock \doi{10.1038/s41598-020-62971-3}.
\newblock URL \url{https://doi.org/10.1038/s41598-020-62971-3}.

\bibitem[Jaynes(1957)]{jaynes1957information}
Edwin~T Jaynes.
\newblock Information theory and statistical mechanics.
\newblock \emph{Physical review}, 106\penalty0 (4):\penalty0 620, 1957.

\bibitem[Karpukhin et~al.(2020)Karpukhin, Oguz, Min, Lewis, Wu, Edunov, Chen, and Yih]{karpukhin-etal-2020-dense}
Vladimir Karpukhin, Barlas Oguz, Sewon Min, Patrick Lewis, Ledell Wu, Sergey Edunov, Danqi Chen, and Wen-tau Yih.
\newblock Dense passage retrieval for open-domain question answering.
\newblock In \emph{Proceedings of the 2020 Conference on Empirical Methods in Natural Language Processing (EMNLP)}, pp.\  6769--6781, Online, November 2020. Association for Computational Linguistics.
\newblock \doi{10.18653/v1/2020.emnlp-main.550}.
\newblock URL \url{https://aclanthology.org/2020.emnlp-main.550}.

\bibitem[Kwiatkowski et~al.(2019)Kwiatkowski, Palomaki, Redfield, Collins, Parikh, Alberti, Epstein, Polosukhin, Devlin, Lee, Toutanova, Jones, Kelcey, Chang, Dai, Uszkoreit, Le, and Petrov]{kwiatkowski-etal-2019-natural}
Tom Kwiatkowski, Jennimaria Palomaki, Olivia Redfield, Michael Collins, Ankur Parikh, Chris Alberti, Danielle Epstein, Illia Polosukhin, Jacob Devlin, Kenton Lee, Kristina Toutanova, Llion Jones, Matthew Kelcey, Ming-Wei Chang, Andrew~M. Dai, Jakob Uszkoreit, Quoc Le, and Slav Petrov.
\newblock Natural questions: A benchmark for question answering research.
\newblock \emph{Transactions of the Association for Computational Linguistics}, 7:\penalty0 452--466, 2019.
\newblock \doi{10.1162/tacl_a_00276}.
\newblock URL \url{https://aclanthology.org/Q19-1026}.

\bibitem[Lin et~al.(2022)Lin, Trivedi, and Sun]{lin2022conformal}
Zhen Lin, Shubhendu Trivedi, and Jimeng Sun.
\newblock Conformal prediction intervals with temporal dependence.
\newblock \emph{Transactions on Machine Learning Research}, 2022.
\newblock ISSN 2835-8856.
\newblock URL \url{https://openreview.net/forum?id=8QoxXTDcsH}.

\bibitem[Müller(1997)]{Muller}
Alfred Müller.
\newblock Integral probability metrics and their generating classes of functions.
\newblock \emph{Advances in Applied Probability}, 29\penalty0 (2):\penalty0 429--443, 1997.
\newblock ISSN 00018678.
\newblock URL \url{http://www.jstor.org/stable/1428011}.

\bibitem[Oliphant(2006)]{numpy}
Travis~E Oliphant.
\newblock \emph{\href{https://web.mit.edu/dvp/Public/numpybook.pdf}{A guide to NumPy}}, volume~1.
\newblock Trelgol Publishing USA, 2006.

\bibitem[Oliveira et~al.(2022)Oliveira, Orenstein, Ramos, and Romano]{oliveira2022split}
Roberto~I. Oliveira, Paulo Orenstein, Thiago Ramos, and João~Vitor Romano.
\newblock Split conformal prediction for dependent data, 2022.

\bibitem[Papadopoulos et~al.(2002)Papadopoulos, Proedrou, Vovk, and Gammerman]{papadopoulos2002inductive}
Harris Papadopoulos, Kostas Proedrou, Volodya Vovk, and Alex Gammerman.
\newblock Inductive confidence machines for regression.
\newblock In Tapio Elomaa, Heikki Mannila, and Hannu Toivonen (eds.), \emph{Machine Learning: ECML 2002}, pp.\  345--356, Berlin, Heidelberg, 2002. Springer Berlin Heidelberg.
\newblock ISBN 978-3-540-36755-0.

\bibitem[Paszke et~al.(2019)Paszke, Gross, Massa, Lerer, Bradbury, Chanan, Killeen, Lin, Gimelshein, Antiga, Desmaison, Kopf, Yang, DeVito, Raison, Tejani, Chilamkurthy, Steiner, Fang, Bai, and Chintala]{pytorch}
Adam Paszke, Sam Gross, Francisco Massa, Adam Lerer, James Bradbury, Gregory Chanan, Trevor Killeen, Zeming Lin, Natalia Gimelshein, Luca Antiga, Alban Desmaison, Andreas Kopf, Edward Yang, Zachary DeVito, Martin Raison, Alykhan Tejani, Sasank Chilamkurthy, Benoit Steiner, Lu~Fang, Junjie Bai, and Soumith Chintala.
\newblock Pytorch: An imperative style, high-performance deep learning library.
\newblock In H.~Wallach, H.~Larochelle, A.~Beygelzimer, F.~d\textquotesingle Alch\'{e}-Buc, E.~Fox, and R.~Garnett (eds.), \emph{Advances in Neural Information Processing Systems 32}, pp.\  8024--8035. Curran Associates, Inc., 2019.

\bibitem[Pedregosa et~al.(2011)Pedregosa, Varoquaux, Gramfort, Michel, Thirion, Grisel, Blondel, Prettenhofer, Weiss, Dubourg, Vanderplas, Passos, Cournapeau, Brucher, Perrot, and Duchesnay]{scikit-learn}
F.~Pedregosa, G.~Varoquaux, A.~Gramfort, V.~Michel, B.~Thirion, O.~Grisel, M.~Blondel, P.~Prettenhofer, R.~Weiss, V.~Dubourg, J.~Vanderplas, A.~Passos, D.~Cournapeau, M.~Brucher, M.~Perrot, and E.~Duchesnay.
\newblock Scikit-learn: Machine learning in {P}ython.
\newblock \emph{Journal of Machine Learning Research}, 12:\penalty0 2825--2830, 2011.

\bibitem[Podkopaev \& Ramdas(2021)Podkopaev and Ramdas]{podkopaev2021label}
Aleksandr Podkopaev and Aaditya Ramdas.
\newblock Distribution-free uncertainty quantification for classification under label shift.
\newblock In Cassio de~Campos and Marloes~H. Maathuis (eds.), \emph{Proceedings of the Thirty-Seventh Conference on Uncertainty in Artificial Intelligence}, volume 161 of \emph{Proceedings of Machine Learning Research}, pp.\  844--853. PMLR, 27--30 Jul 2021.
\newblock URL \url{https://proceedings.mlr.press/v161/podkopaev21a.html}.

\bibitem[Reimers \& Gurevych(2019)Reimers and Gurevych]{reimers-gurevych-2019-sentence}
Nils Reimers and Iryna Gurevych.
\newblock Sentence-{BERT}: Sentence embeddings using {S}iamese {BERT}-networks.
\newblock In \emph{Proceedings of the 2019 Conference on Empirical Methods in Natural Language Processing and the 9th International Joint Conference on Natural Language Processing (EMNLP-IJCNLP)}, pp.\  3982--3992, Hong Kong, China, November 2019. Association for Computational Linguistics.
\newblock \doi{10.18653/v1/D19-1410}.
\newblock URL \url{https://aclanthology.org/D19-1410}.

\bibitem[Saunders et~al.(1999)Saunders, Gammerman, and Vovk]{saunders1999transduction}
Craig Saunders, Alexander Gammerman, and Volodya Vovk.
\newblock Transduction with confidence and credibility.
\newblock In \emph{Proceedings of the Sixteenth International Joint Conference on Artificial Intelligence}, IJCAI '99, pp.\  722–726, San Francisco, CA, USA, 1999. Morgan Kaufmann Publishers Inc.
\newblock ISBN 1558606130.

\bibitem[Schlembach et~al.(2022)Schlembach, Smirnov, and Koprinska]{schlembach2022conformal}
Filip Schlembach, Evgueni Smirnov, and Irena Koprinska.
\newblock Conformal multistep-ahead multivariate time-series forecasting.
\newblock In Ulf Johansson, Henrik Boström, Khuong An~Nguyen, Zhiyuan Luo, and Lars Carlsson (eds.), \emph{Proceedings of the Eleventh Symposium on Conformal and Probabilistic Prediction with Applications}, volume 179 of \emph{Proceedings of Machine Learning Research}, pp.\  316--318. PMLR, 24--26 Aug 2022.
\newblock URL \url{https://proceedings.mlr.press/v179/schlembach22a.html}.

\bibitem[Sch{\"o}lkopf et~al.(1998)Sch{\"o}lkopf, Bartlett, Smola, and Williamson]{scholkopf1998shrinking}
Bernhard Sch{\"o}lkopf, Peter Bartlett, Alex Smola, and Robert~C Williamson.
\newblock Shrinking the tube: a new support vector regression algorithm.
\newblock \emph{Advances in neural information processing systems}, 11, 1998.

\bibitem[Schuster et~al.(2022)Schuster, Fisch, Gupta, Dehghani, Bahri, Tran, Tay, and Metzler]{schuster2022confident}
Tal Schuster, Adam Fisch, Jai Gupta, Mostafa Dehghani, Dara Bahri, Vinh~Q. Tran, Yi~Tay, and Donald Metzler.
\newblock Confident adaptive language modeling.
\newblock In Alice~H. Oh, Alekh Agarwal, Danielle Belgrave, and Kyunghyun Cho (eds.), \emph{Advances in Neural Information Processing Systems}, 2022.
\newblock URL \url{https://openreview.net/forum?id=uLYc4L3C81A}.

\bibitem[Stankeviciute et~al.(2021)Stankeviciute, M~Alaa, and van~der Schaar]{stankeviciute2021conformal}
Kamile Stankeviciute, Ahmed M~Alaa, and Mihaela van~der Schaar.
\newblock Conformal time-series forecasting.
\newblock \emph{Advances in neural information processing systems}, 34:\penalty0 6216--6228, 2021.

\bibitem[Sun \& Yu(2022)Sun and Yu]{sun2022copula}
Sophia Sun and Rose Yu.
\newblock Copula conformal prediction for multi-step time series forecasting, 2022.

\bibitem[Teneggi et~al.(2023)Teneggi, Tivnan, Stayman, and Sulam]{teneggi2023trust}
Jacopo Teneggi, Matthew Tivnan, Web Stayman, and Jeremias Sulam.
\newblock How to trust your diffusion model: A convex optimization approach to conformal risk control.
\newblock In Andreas Krause, Emma Brunskill, Kyunghyun Cho, Barbara Engelhardt, Sivan Sabato, and Jonathan Scarlett (eds.), \emph{Proceedings of the 40th International Conference on Machine Learning}, volume 202 of \emph{Proceedings of Machine Learning Research}, pp.\  33940--33960. PMLR, 23--29 Jul 2023.
\newblock URL \url{https://proceedings.mlr.press/v202/teneggi23a.html}.

\bibitem[Tibshirani et~al.(2019)Tibshirani, Foygel~Barber, Candes, and Ramdas]{tibshirani2019covariate}
Ryan~J Tibshirani, Rina Foygel~Barber, Emmanuel Candes, and Aaditya Ramdas.
\newblock Conformal prediction under covariate shift.
\newblock In H.~Wallach, H.~Larochelle, A.~Beygelzimer, F.~d\textquotesingle Alch\'{e}-Buc, E.~Fox, and R.~Garnett (eds.), \emph{Advances in Neural Information Processing Systems}, volume~32. Curran Associates, Inc., 2019.
\newblock URL \url{https://proceedings.neurips.cc/paper_files/paper/2019/file/8fb21ee7a2207526da55a679f0332de2-Paper.pdf}.

\bibitem[Van~Rossum \& Drake(2009)Van~Rossum and Drake]{python}
Guido Van~Rossum and Fred~L. Drake.
\newblock \emph{\href{https://dl.acm.org/doi/book/10.5555/1593511}{Python 3 Reference Manual}}.
\newblock CreateSpace, Scotts Valley, CA, 2009.
\newblock ISBN 1441412697.

\bibitem[Vapnik(1999)]{vapnik1999nature}
Vladimir Vapnik.
\newblock \emph{The nature of statistical learning theory}.
\newblock Springer science \& business media, 1999.

\bibitem[{Virtanen} et~al.(2020){Virtanen}, {Gommers}, {Oliphant}, {Haberland}, {Reddy}, {Cournapeau}, {Burovski}, {Peterson}, {Weckesser}, {Bright}, {van der Walt}, {Brett}, {Wilson}, {Jarrod Millman}, {Mayorov}, {Nelson}, {Jones}, {Kern}, {Larson}, {Carey}, {Polat}, {Feng}, {Moore}, {Vand erPlas}, {Laxalde}, {Perktold}, {Cimrman}, {Henriksen}, {Quintero}, {Harris}, {Archibald}, {Ribeiro}, {Pedregosa}, {van Mulbregt}, and {Contributors}]{scipy}
Pauli {Virtanen}, Ralf {Gommers}, Travis~E. {Oliphant}, Matt {Haberland}, Tyler {Reddy}, David {Cournapeau}, Evgeni {Burovski}, Pearu {Peterson}, Warren {Weckesser}, Jonathan {Bright}, St{\'e}fan~J. {van der Walt}, Matthew {Brett}, Joshua {Wilson}, K.~{Jarrod Millman}, Nikolay {Mayorov}, Andrew R.~J. {Nelson}, Eric {Jones}, Robert {Kern}, Eric {Larson}, CJ~{Carey}, {\.I}lhan {Polat}, Yu~{Feng}, Eric~W. {Moore}, Jake {Vand erPlas}, Denis {Laxalde}, Josef {Perktold}, Robert {Cimrman}, Ian {Henriksen}, E.~A. {Quintero}, Charles~R {Harris}, Anne~M. {Archibald}, Ant{\^o}nio~H. {Ribeiro}, Fabian {Pedregosa}, Paul {van Mulbregt}, and SciPy 1.~0 {Contributors}.
\newblock \href{https://doi.org/10.1038/s41592-019-0686-2}{SciPy 1.0: Fundamental Algorithms for Scientific Computing in Python}.
\newblock \emph{Nature Methods}, 2020.
\newblock \doi{https://doi.org/10.1038/s41592-019-0686-2}.

\bibitem[Vovk(2015)]{vovk2015cross}
Vladimir Vovk.
\newblock Cross-conformal predictors.
\newblock \emph{Annals of Mathematics and Artificial Intelligence}, 74\penalty0 (1):\penalty0 9--28, 2015.
\newblock \doi{10.1007/s10472-013-9368-4}.
\newblock URL \url{https://doi.org/10.1007/s10472-013-9368-4}.

\bibitem[Vovk et~al.(2005)Vovk, Gammerman, and Shafer]{vovk2005algorithmic}
Vladimir Vovk, Alex Gammerman, and Glenn Shafer.
\newblock \emph{Algorithmic Learning in a Random World}.
\newblock Springer-Verlag, Berlin, Heidelberg, 2005.
\newblock ISBN 0387001522.

\bibitem[Vovk et~al.(1999)Vovk, Gammerman, and Saunders]{vovk1999machine}
Volodya Vovk, Alexander Gammerman, and Craig Saunders.
\newblock Machine-learning applications of algorithmic randomness.
\newblock In \emph{Proceedings of the Sixteenth International Conference on Machine Learning}, ICML '99, pp.\  444–453, San Francisco, CA, USA, 1999. Morgan Kaufmann Publishers Inc.
\newblock ISBN 1558606122.

\bibitem[Walt et~al.(2011)Walt, Colbert, and Varoquaux]{nparray}
St{\'e}fan van~der Walt, S~Chris Colbert, and Gael Varoquaux.
\newblock \href{https://arxiv.org/abs/1102.1523}{The NumPy array: a structure for efficient numerical computation}.
\newblock \emph{Computing in Science \& Engineering}, 13\penalty0 (2):\penalty0 22--30, 2011.

\bibitem[Wang et~al.(2023)Wang, Tong, Tan, Vorobeychik, and Kantaros]{wang2023conformal}
Jun Wang, Jiaming Tong, Kaiyuan Tan, Yevgeniy Vorobeychik, and Yiannis Kantaros.
\newblock Conformal temporal logic planning using large language models: Knowing when to do what and when to ask for help, 2023.

\bibitem[Xu \& Xie(2021)Xu and Xie]{xu2021conformal}
Chen Xu and Yao Xie.
\newblock Conformal prediction interval for dynamic time-series.
\newblock In Marina Meila and Tong Zhang (eds.), \emph{Proceedings of the 38th International Conference on Machine Learning}, volume 139 of \emph{Proceedings of Machine Learning Research}, pp.\  11559--11569. PMLR, 18--24 Jul 2021.
\newblock URL \url{https://proceedings.mlr.press/v139/xu21h.html}.

\bibitem[Yang \& Pavone(2023)Yang and Pavone]{yang2023object}
Heng Yang and Marco Pavone.
\newblock Object pose estimation with statistical guarantees: Conformal keypoint detection and geometric uncertainty propagation.
\newblock \emph{2023 IEEE/CVF Conference on Computer Vision and Pattern Recognition (CVPR)}, Jun 2023.
\newblock \doi{10.1109/cvpr52729.2023.00864}.
\newblock URL \url{http://dx.doi.org/10.1109/CVPR52729.2023.00864}.

\bibitem[Zaffran et~al.(2022)Zaffran, Feron, Goude, Josse, and Dieuleveut]{zaffran2022adaptive}
Margaux Zaffran, Olivier Feron, Yannig Goude, Julie Josse, and Aymeric Dieuleveut.
\newblock Adaptive conformal predictions for time series.
\newblock In Kamalika Chaudhuri, Stefanie Jegelka, Le~Song, Csaba Szepesvari, Gang Niu, and Sivan Sabato (eds.), \emph{Proceedings of the 39th International Conference on Machine Learning}, volume 162 of \emph{Proceedings of Machine Learning Research}, pp.\  25834--25866. PMLR, 17--23 Jul 2022.
\newblock URL \url{https://proceedings.mlr.press/v162/zaffran22a.html}.

\end{thebibliography}
\bibliographystyle{iclr2024_conference}

\appendix
\section{Proof of \cref{lemma:tv_bound}}\label{sec:proof_lemma}

The TV distance can be written as an integral probability metric  \citep{Muller}:

\begin{equation}\label{eq:ipm}
    d_{\mathrm{TV}}(P, Q) = \frac{1}{2}\sup_{g:\, \| g \|_{\infty}\leq 1} \bigl( \mathbb{E}_P[g] - \mathbb{E}_Q[g]\bigr).
\end{equation}

Now, we define $m = (A+B) / 2$, $v = (B - A) / 2$, and $\bar{f} = (f - m)/v:S \rightarrow [-1,\, 1] $. 
Noticing that for any $c\in\mathbb{R}$, we have $\mathbb{E}_P[f] - \mathbb{E}_Q[f] = \mathbb{E}_P[f+c] - \mathbb{E}_Q[f+c]$, we can evaluate the difference in expectations as 

\begin{align}
    \mathbb{E}_P[f] - \mathbb{E}_Q[f] & = v\, \left(\mathbb{E}_P[\bar{f}] - \mathbb{E}_Q[\bar{f}]\right) \\
    & \leq \frac{B-A}{2}\sup_{g:\, \| g \|_{\infty}\leq 1} \bigl( \mathbb{E}_P[g] - \mathbb{E}_Q[g]\bigr) \\
    & = (B-A) \; d_{\mathrm{TV}}(P, Q).
\end{align}

Repeating with $\bar{f}=(m-f)/v$ (which is also in $[-1,\,1]$), yields a similar upper-bound for $\mathbb{E}_Q[f] - \mathbb{E}_P[f] $, from which the result for $|\mathbb{E}_P[f] - \mathbb{E}_Q[f]|$ follows.

\section{Proof of \cref{theorem:nonx-crc}}\label{sec:proof_theorem}

The proof adapts elements of the proofs from \cite{barber2022conformal} and \cite{angelopoulos2022conformal}. 
Let $Z^K$ be obtained from $Z$ by swapping $(X_K, Y_K)$ and $(X_{n+1}, Y_{n+1})$, where $K$ is a random variable where $\mathbb{P}\{K = i\} = \tilde{w}_i$ (note that $Z^{n+1} = Z$). 
Let 
\begin{align}
    \hat{R}_{n+1}(\lambda) = \sum_{i=1}^{n+1} \tilde{w}_i L(\lambda; (x_i, y_i)) = 
    \frac{N_w \hat{R}_n(\lambda) + L(\lambda; (x_{n+1}, y_{n+1}))}{N_w+1}
\end{align}
be the weighted empirical risk in the calibration set plus the additional test example. 
Let us define
\begin{align}\label{eq:lambda_star}
\lambda^* &= \mathrm{inf} \left\{ \lambda : \hat{R}_{n+1}(\lambda)  \le \alpha \right\}.
\end{align}

Given the random variable $Z$, 
we can think of $\lambda^*(Z)$ as another random variable which is a transformation of $Z$. 
Moreover, we define the random variable $F_i(Z) = L(\lambda^*(Z); (X_i, Y_i))$ for $i \in [n+1]$, 
as well as the vector of random variables $F(Z) = [F_1(Z), \ldots, F_{n+1}(Z)]$. 
From \cref{lemma:tv_bound}, we have 
\begin{align}
    \mathbb{E}[F_i(Z^i)] \,\,\le\,\, & \mathbb{E}[F_i(Z)] + (B-A)d_{\mathrm{TV}}(F(Z), F(Z^i)),
\end{align}
a bound that we will use later.
Writing $L_i(\lambda) \equiv L(\lambda; (X_i, Y_i))$ for convenience, we also have, for any $\lambda$ and for any $k \in [n+1]$, 
\begin{align}
\hat{R}_{n+1}(\lambda; Z^k) \,\,&=\,\, \sum_{i=1, i \ne k}^{n} \tilde{w}_i L_i(\lambda) +  \tilde{w}_k L_{n+1}(\lambda) + \tilde{w}_{n+1} L_k(\lambda)\nonumber\\
\,\,&=\,\, \sum_{i=1, i \ne k}^{n} \tilde{w}_i L_i(\lambda) +  \tilde{w}_k (L_k(\lambda) + \underbrace{L_{n+1}(\lambda)}_{\le B}) + \underbrace{(\tilde{w}_{n+1} - \tilde{w_k})}_{\ge 0} \underbrace{L_k(\lambda)}_{\le B}\nonumber\\
\,\,&\le \,\, \sum_{i=1, i \ne k}^{n} \tilde{w}_i L_i(\lambda) +  \tilde{w}_k (L_{k}(\lambda) + B) + (\tilde{w}_{n+1} - \tilde{w}_k) B\nonumber\\
\,\,&= \,\, \sum_{i=1}^{n} \tilde{w}_i L_i(\lambda) +  \tilde{w}_{n+1} B \nonumber\\
\,\,&= \,\, 
\frac{N_w}{N_w + 1} \hat{R}_n(\lambda; Z) + \frac{B}{N_w+1}.
\end{align}
Therefore, setting $\lambda = \hat{\lambda}$ and using \cref{eq:lambda_hat}, we obtain
    $\hat{R}_{n+1}(\hat{\lambda}; Z^k) \,\le \, \frac{N_w}{N_w + 1} \hat{R}_n(\hat{\lambda}; Z) + \frac{B}{N_w+1}
    \,\le\, \alpha$,
which, from \cref{eq:lambda_star}, implies $\lambda^*(Z^k) \le \hat{\lambda}(Z)$. 
Since the loss $L$ is nonincreasing with $\lambda$, we get
\begin{align}\label{eq:bound}
    \mathbb{E}[L_{n+1}(\hat{\lambda}(Z); Z)] \,\,&\le\,\, \mathbb{E}[L_{n+1}(\lambda^*(Z^K); Z)]
    \,\,=\,\, \mathbb{E}[L_K(\lambda^*(Z^K); Z^K] \nonumber\\
    \,\,&=\,\, \sum_{i=1}^{n+1} \underbrace{\mathbb{P}\{K=i\}}_{=\tilde{w}_i} \underbrace{\mathbb{E}[L_i(\lambda^*(Z^i), Z^i]}_{= \mathbb{E}[F_i(Z^i)]} \nonumber\\
    \,\,&\le\,\, \sum_{i=1}^{n+1} \tilde{w}_i \left(\underbrace{ \mathbb{E}[L_i(\lambda^*(Z), Z]}_{= \mathbb{E}[F_i(Z)]} + (B-A)d_{\mathrm{TV}}(F(Z), F(Z^i)) \right)  \nonumber\\  
    \,\,&=\,\, \mathbb{E}\left[\sum_{i=1}^{n+1} \tilde{w}_i L_i(\lambda^*(Z), Z)\right] + (B-A)\sum_{i=1}^n \tilde{w}_i d_{\mathrm{TV}}(F(Z), F(Z^i))   
    \nonumber\\   
    \,\,&=\,\, \mathbb{E}\left[\hat{R}_{n+1}(\lambda^*(Z))\right] + (B-A)\sum_{i=1}^n \tilde{w}_i d_{\mathrm{TV}}(F(Z), F(Z^i))\nonumber\\
    \,\,&\le\,\, \alpha + (B-A)\sum_{i=1}^n \tilde{w}_i d_{\mathrm{TV}}(F(Z), F(Z^i)).    
\end{align}
The result follows by noting that $d_{\mathrm{TV}}(F(Z), F(Z^i)) \le d_{\mathrm{TV}}(Z, Z^i)$.
\cref{eq:bound} is actually a tighter bound, similarly to what has been noted by \citealp{barber2022conformal}.

\section{Multilabel classification in a time series}
\label{app:experiment-synthetic}

\begin{figure}[t]
\begin{center}
\includegraphics[width=1.0\columnwidth]{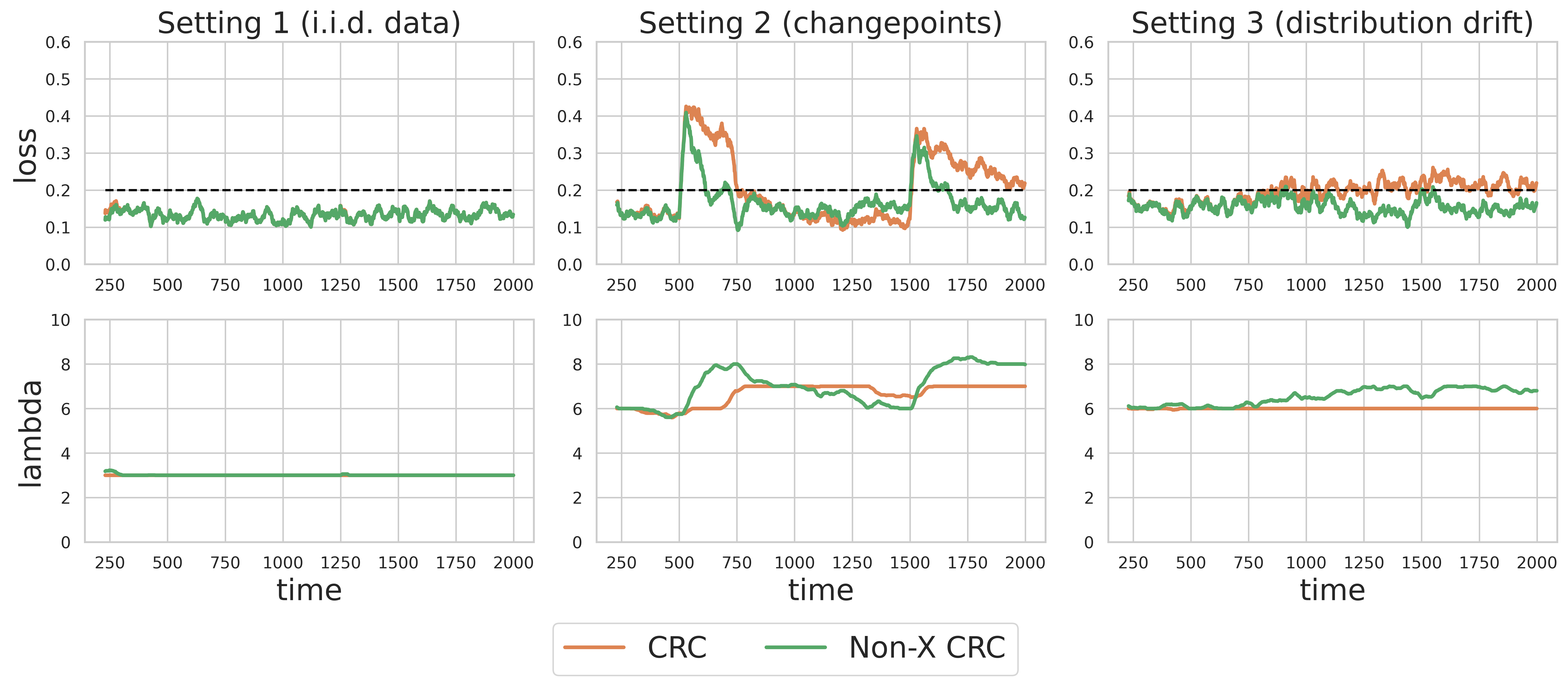}
\caption{Average loss (top) and $\hat{\lambda}$ (bottom) over $10$ independent trials for settings (1), (2), and (3). In this case, $\lambda$ represents the number of predicted labels. We smooth the curves by taking a rolling average with a window of $30$ time points.
\label{fig:plot-simulations-numberelements}}
\end{center}
\end{figure}

\cref{fig:plot-simulations-numberelements} shows results averaged across $10$ independent trials for $\alpha=0.2$ and setting $\lambda$ in a slightly different way than that of \cref{sec:experiment-synthetic}.
In this case, $\lambda$ represents the number of active labels and we use $\mathcal{C}_\lambda(X_i) = \text{top-$\lambda$}(\bm{f}(X_i))$.
The main takeaways remain the same: both methods perform similarly when the data is exchangeable, in setting (1). Accounting for the non-exchangeability introduced by changepoints and distribution drift using our method enables lowering the risk to the desired level in settings (2) and (3).

\end{document}